\documentclass[]{interact}

\usepackage{graphicx}
\usepackage{apacite}
\usepackage[]{natbib}
\usepackage{color}

\usepackage[caption=false]{subfig}

\theoremstyle{plain}

\usepackage{longtable}
\usepackage{multirow}
\theoremstyle{definition}

\usepackage{multirow}
\usepackage{hyperref}
\hypersetup{hypertex=true,
            colorlinks=true,
            linkcolor=blue,
            anchorcolor=blue,
            citecolor=blue}
\theoremstyle{remark}

\renewcommand{\abstractname}{\textcolor{blue}{ABSTRACT} \\}
\renewcommand{\keywordsname}{\textcolor{blue}{KEYWORDS} \\}
\renewcommand{\figurename}{\textcolor{blue}{Figure} \\}

\begin{document}

\articletype{ARTICLE TEMPLATE}

\title{Applications of Federated Learning in Smart Cities: Recent Advances, Taxonomy, and Open Challenges }

\author{
\name{Zhaohua Zheng\textsuperscript{1,2}\thanks{CONTACT Zhaohua Zheng, Email: zhengzhaohua@tju.edu.cn, \\Zhaohua Zheng and Yize Zhou contribute equally to this work.\\},Yize Zhou\textsuperscript{2},Yilong Sun\textsuperscript{2},Zhang Wang\textsuperscript{2}, Boyi Liu\textsuperscript{3} and Keqiu Li\textsuperscript{1}}
\affil{\textsuperscript{1}Tianjin University, China;\textsuperscript{2}Hainan University, China;\textsuperscript{3}University of Macau}
}

\maketitle

\begin{abstract}
Federated learning (FL) plays an important role in the development of smart cities. With the development of big data and artificial intelligence, there is a problem of data privacy protection and FL can solve this problem. This study reviews the current developments in FL and its applications in various fields. A comprehensive investigation is performed, and the latest research on the application of FL is discussed for various fields of smart cities. We explain in depth current developments of FL in fields such as the Internet of Things (IoT), transportation, communications, finance, and medicine. First, we introduce the background, definition, and key technologies of FL. Then, we review key 
applications and the latest results. Finally, we discuss the future applications and research directions of FL in smart cities.
\end{abstract}


\textcolor{blue}{\section{Introduction}}

\noindent A smart city refers to the application of different types of electronic IOT sensors to collect data (Mahmud et al., \citeyear{mahmud2018internet}) in urban environments. Although existing urban Internet architecture is complex, decision makers can use the insights gained from this data to effectively manage the assets, resources, and services in urban areas. The operating model of a smart city is to use IoT sensors to collect data. Furthermore, it can realize effective applications in a series of fields, such as urban public services, resource allocation, and communications. Simultaneously, smart cities have provided good solutions to key issues such as the development of the IoT (Li, Zhao, \& Wong, \citeyear{li2020machine}),medical care (Rath \& Pattanayak, \citeyear{rath2019technological}), transportation (Qiu et al., \citeyear{qiu2019nei}), communications (Guan et al., \citeyear{guan2018privacy}), etc. In this large-scale information exchange process, sensors generate a large amount of data. These data are of great significance for improving the application of the program and helping managers optimize their decisions. However, a large proportion of the data was sensitive, and it involves user-generated personal privacy (Brisimi et al., \citeyear{khan2019federated}). First, we must prevent the inclusion of users' private data during data processing. Additionally, there are the problems of low data resource utilization and network congestion (Khan et al., \citeyear{mcmahan2017communication}) in the process of data interaction.

At present, self-organization theory (J. Yan et al., \citeyear{yan2020evaluation}), machine learning (ML) (Li, Zhao,
\& Wong, \citeyear{li2020machine}), edge computing nodes (Rahman et al., \citeyear{rahman2019blockchain}), system simulation (Lv et al., \citeyear{lv2019infrastructure}) and other computing implementations all have large network bottlenecks in practical applications related to smart cities. In addition, it still has the problem of low efficiency when using network resources, and a distributed learning paradigm is therefore required. Distributed framework can reduce network bottlenecks. The problem of user privacy is solved through the collaborative sharing model of IoT devices. Some examples of distributed organization computing are mentioned in current smart city applications. However, these practical applications have not adequately solved the problem of user privacy (Zhou et al., \citeyear{zhou2017greening}).

\begin{figure}
\centering
\includegraphics[width=1\linewidth]{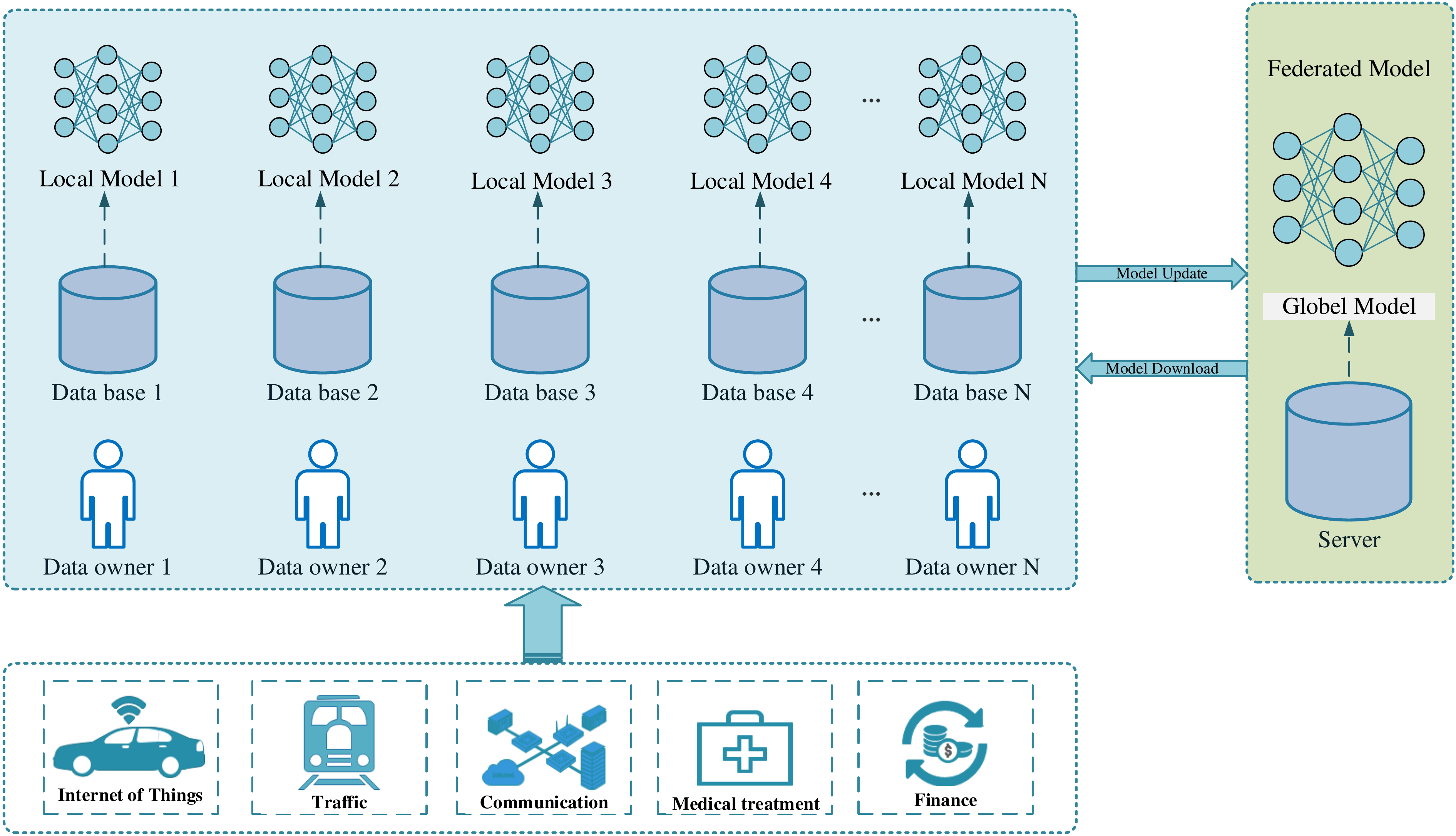}
\caption{Application of Federated Learning in Smart Cities.}
\label{fig:architecture}
\end{figure} 

FL has the advantage of solving the aforementioned problems. Under the FL framework, users can make use of the data without obtaining other participants' data. The related data are stored locally (B. McMahan et al., \citeyear{mcmahan2017communication}). Users periodically share their local model gradients only with the coordination server for a period of time. The server organizes the training data and measures the contributions of all of the participants (Smith, Forte, et al., \citeyear{16smith2017cocoa}). It constructs a global model by averaging all gradients in the network (Li et al., \citeyear{17li2018federated}) at the server level. Subsequently, the coordination server distributes the new model update to all clients (Lim, Luong, et al.,  \citeyear{18lim2020federated}). Each client uploads its local model to the server. Then, users download new updated models and use cloud distribution to realize inference on the device. The coordination process of the entire server continues until it stops. This is the complete operating principle of FL algorithms.

FL has the advantages of distributed processing and effective privacy protection. Some common distributed communication devices, such as mobile phones, have communication transfer problems. FL proposes a federated domain-adaptive method that is based on the domain-transfer problem. This model solves the problem of data privacy and efficiency  (Peng et al., \citeyear{20peng2019federated}). Meanwhile, some scholars have implemented block-chain FL (BlockFL) architecture that can realize the exchange and verification of local learning model updates. It can describe the best block generation rate by considering communication and consensus delay issues (Kim et al., \citeyear{21kim2019blockchained}). Related research results related to FL have been conducted in the fields of IoT, communications, and public services. These practices promote the updating and development of applications in smart cities. In Figure \ref{fig:architecture}, we show the application of FL in smart city related fields.

The remainder of this paper is organized as follows. Section 2 introduces the definition and key technologies of FL, and  section 3 explains the applications of FL in the IoT system of smart cities. Then, section 4 discusses the applications of FL in transportation systems. Section 5 presents the applications of FL in the financial field of smart cities, and section 6 introduces the applications of FL to the medical field of smart cities. Section 7 explains the communication of FL in the field of smart cities. Then, section 8 discusses the future developments and directions of FL in smart cities. Finally, the conclusions are presented.

\begin{figure}
\centering
\includegraphics[width=1\linewidth]{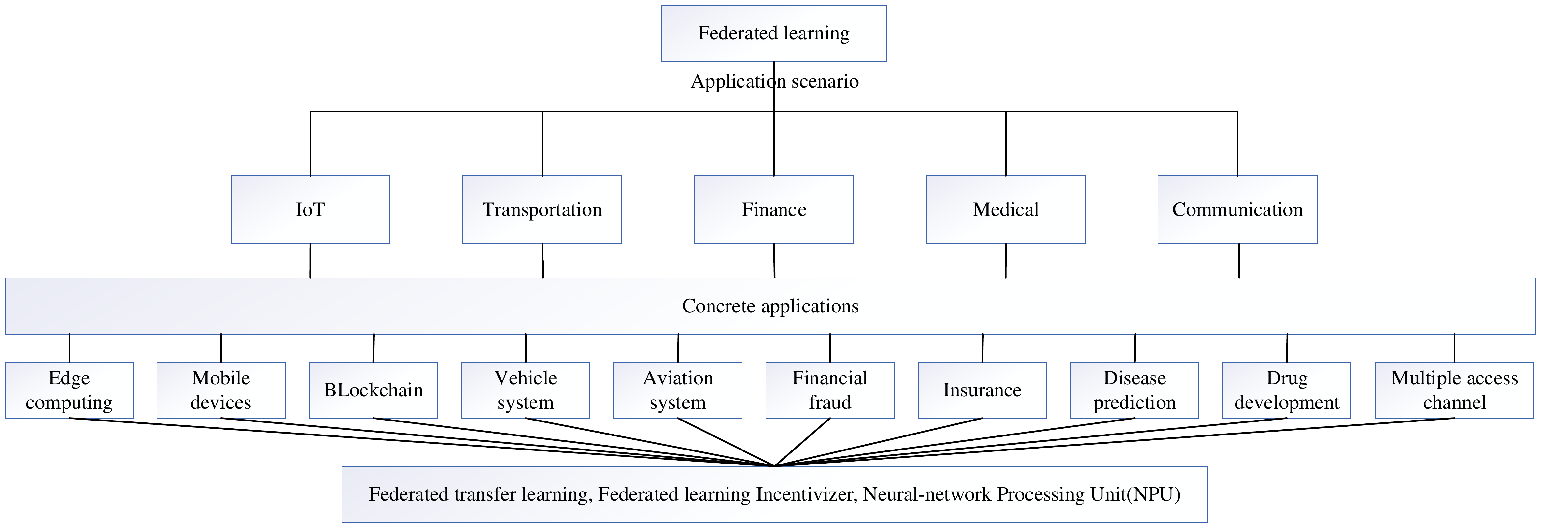}
\caption{Federated learning involved fields and methods in Smart Cities.}
\label{fig:fields and methods}
\end{figure}

Up to the present, FL is not considered very popular in smart city development, and this led us to conduct a comprehensive investigation. This study makes the following contributions.
\begin{itemize}
\item This work introduces the background, definition, and key technologies of FL.
\item This work classifies and summarizes the latest research on the application of FL in smart cities. Simultaneously, we review key technologies and the latest results in Figure \ref{fig:fields and methods}.
\item We discuss the future development of applications and research directions of FL in smart cities.
\end{itemize}

\textcolor{blue}{\section{Definition and key technologies of federated learning}}

\noindent The concept of FL has been widely proposed (Kone\u cn\`y, McMahan, Ramage, \&
Richt\'arik, \citeyear{23konevcny2016federated}), implemented, and applied in various fields. Most of the existing large-scale studies have been applied to distributed learning in the development of big data (Dayarathna
et al., \citeyear{dayarathna2017role}) and cloud computing (Dean et al., \citeyear{27dean2012large}). The core of FL is to build a ML model using data distributed across multiple devices. This solves the problem of data privacy. Currently, there is a rapid increase in the use of distributed computing agents, and FL has become an effective solution to this problem as it protects user privacy in the process of information and knowledge sharing (Smith, Chiang, et al., \citeyear{19smith2017federated}). For example, smart behaviors currently exist in mobile devices. Mobile phones and tablets use image classification to predict the classification of pictures that have been previewed multiple times (Ji et al., \citeyear{29ji2019learning}). FL is based on the use of data and information processing to improve user experience. In addition, many insurance companies have always been concerned about protecting their data, which they are unwilling to share with other entities (G. Wang et al., \citeyear{wang2019measure}). In this case, multiparty data can be used in the FL framework as it solves the privacy problem present in ML. Recent research improvements in FL have mainly focused on statistical challenges in FL (Smith, Chiang, et al., \citeyear{19smith2017federated}) and security issues (Bonawitz et al., \citeyear{32bonawitz2017practical}). At the same time, research has made FL more personalized (F. Chen et al., \citeyear{34chen2018federated}). This process involves factors such as data interaction among distributed mobile users, unbalanced data distribution, and communication costs in equipment reliability. It can inspire researchers to continuously overcome challenges that are related to data privacy, computational constraints, and communication costs. In addition, the concept of FL is extended to include other collaborative learning programs between organizations. Here, we provide a preliminary explanation of the extension of the original concept of FL to other distributed collaborative ML. Then, in this article, we further investigate the application of FL in smart cities. We also discuss its development status and future directions. In this section, we provide a more comprehensive overview of FL, and consider the definition, privacy, training process, and classification structure of FL.

\textcolor{blue}{\subsection{Basic definition of FL}}

\noindent We define $N$ data owners as ${F_{1},F_{2},...,F_{N}}$, all of whom hope to train their own ML models by merging their respective data ${D_{1},D_{2},...,D_{N}}$. A conventional method is to combine all the data. It uses $D=D_{1} \cup D_{2}...\cup D_{N}$ to train and obtain the model $M_{SUM}$. Federated learning is a systematic learning process in which the owners of the data jointly train the model $M_{FED}$. Data owner $F_{i}$ will not disclose his own data $D_{i}$ to others. In addition, the accuracy of $M_{FED}$, which is expressed as $V_{FED}$, should be very close to the performance of $V_{SUM}$ of $M_{SUM}$. In terms of expression, let $\varepsilon$ be a non-negative real number; if $\mid V_{FED}-V_{SUM}\mid < \varepsilon$, then we can assume that the FL algorithm has an $\varepsilon$ error accuracy.

\textcolor{blue}{\subsection{Privacy technologies of FL}}

\noindent Privacy management is one of the core elements considered in FL. The realization of this requirement requires the analyses of security models. In this section, we briefly describe the different privacy technologies that are currently used for FL.

\textcolor{blue}{\subsubsection{Secure Multi-party Computation (SMC) model}}

\noindent SMC model involves data from multiple parties, and it provides safety certification under a known and clear simulation framework. This model guarantees zero interaction of the knowledge data. In this case, in addition to the input and output terminals, neither user knows these information data. The zero-knowledge model formed under this condition is highly expected. Based on this feature, we can consider part of the public knowledge under the complex and secure computing protocol. At present, research has shown that SMC can be used to establish a security model to improve computing efficiency under low security conditions. In addition, the multiparty computation (MPC) protocol performs model training and verification. In this process, users do not need to disclose privacy-sensitive data (Kilbertus et al., \citeyear{37kilbertus2018blind}). However, SMC still has its weaknesses. First of all, as an algorithm for data privacy protection, it cannot deal with the curiosity from the server itself. And for privacy attacks from other clients, the FL system itself has already an acceptable defense. Besides of this, because SMC is a four-round interactive protocol, the server does not learn the client data before completing the submission phase, which will undoubtedly cause a certain amount of data waste and reduce the model accuracy.

\textcolor{blue}{\subsubsection{Differential privacy}}

\noindent Existing research uses differential privacy or technology to achieve the protection of data privacy (McMahan et al., \citeyear{40mcmahan2017learning}). The above methods complete the data processing, and they achieve the purpose of masking certain privacy-sensitive attributes. This makes it impossible for third-party users to distinguish between users, and will make the data unrecoverable and realize the protection of user privacy. However, the disadvantage of this method is that the data must be transferred to other locations, which may affect the accuracy of the data. Therefore, we need to make a trade-off between accuracy and privacy. Currently, many applications have implemented this privacy-processing method. Some researchers have proposed a differential privacy method for FL. They have been enabled to hide customer contributions during the training process to protect client data (Geyer et al., \citeyear{33geyer2017differentially}).

\textcolor{blue}{\subsubsection{Homomorphic encryption}}

\noindent The homomorphic encryption operation model is an encryption mechanism in the machine-learning process. It uses parameter exchange to protect the privacy of user data (Giacomelli et al., \citeyear{43giacomelli2018privacy}). The difference between homomorphic encryption and differential privacy protection is that the data and models themselves will not be transmitted. Their data are also encrypted without discovery. Therefore, its advantage is that the probability of leakage in the original data is very small. In practice, the additive homomorphic encryption model (Acar et al., \citeyear{45acar2018survey}) is widely used.

\textcolor{blue}{\subsection{Typical architecture and training process of FL system}}

\noindent In the FL training system, the owner of the data participates in the FL system. They train a shared model in the aggregation server center. In this architecture, the basic premise is honest data owners and accurate data. This requires data users to employ real private data for training. After the training is completed, the relevant parameters of the local model training are submitted to the FL server.

Generally, the FL training process includes the three training steps mentioned below. We first define the local model as the model trained on each participating device. The global model refers to the model after the FL server is aggregated.

\begin{itemize}
\item \textbf {Step 1.} Initialize the task.  The server determines the training task. In other words, the target application and corresponding data requirements are determined. At the same time, the server specifies the global model and establishes the parameters in the training process, such as the learning rate. The global model parameter $W_{G}^{0}$ is initialized by the server. The training tasks were assigned to the participating users to complete the task assignment.
\item \textbf {Step 2.} Perform training and updating of the local model.  The training was performed on the basis of the global model $W_{G}^{t}$. Here, $t$ represents the current iteration index. Each participating client uses local data and equipment to update the local model parameter $W_{i}^{t}$. The ultimate goal of participant $i$ in iteration $t$ is to find the optimal parameter $W_{i}^{t}$ that minimizes the loss function $L(W_{i}^{t})$, namely $W_{i}^{t^{*}}=\arg min W_{i}^{t}$.
\item \textbf {Step 3.} Realize the aggregation and updating of the global model.  The server aggregates the local models of participating users, and it sends the updated global model parameter $W_{G}^{t+1}$ to the user who holds the data. The server continuously calculates the minimum global loss function $L(W_{G}^{t})$, that is, $L(W_{G}^{t})=\frac{1}{N}\sum_{i=1}^N W_{i}^{t}$ Repeat steps 2-3 until the training global loss function converges or reaches the required training accuracy.
\end{itemize}

\textcolor{blue}{\section{Application of federated learning to IoT system in smart city}}

\begin{figure}[thpb]
	\centering
	\includegraphics[width=1\linewidth]{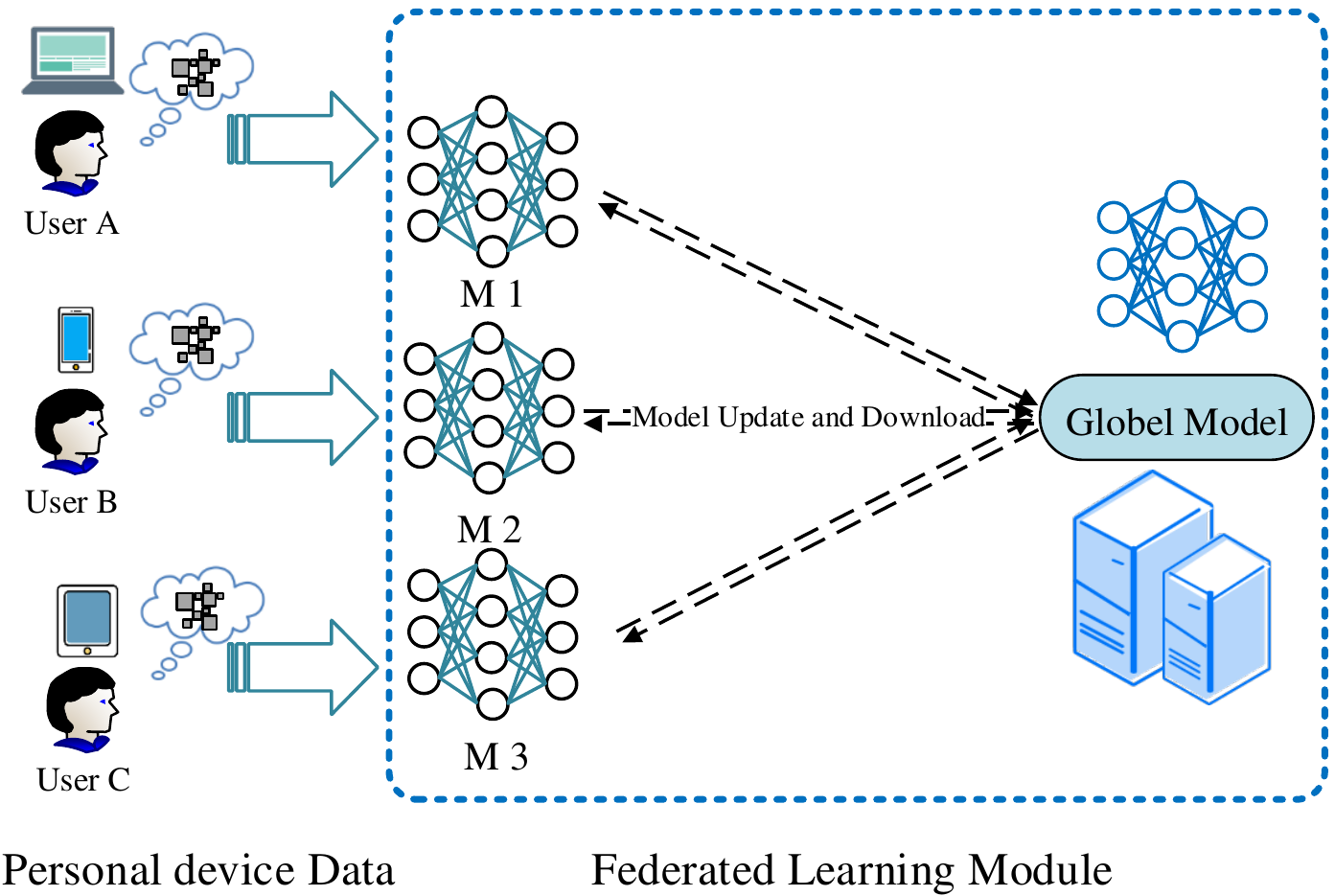}
	\caption{Framework of FL training model for data transmission of multiple mobile devices in IoT}
	\label{fig:Framework}
\end{figure}

\noindent The development of IoT and the application of FL has enabled the provision of technical support for the transformation and progress of smart cities. However, many user privacy and information security issues have been exposed. The framework model of “FL + IoT” has solved many problems. FL builds a scalable production system for mobile devices (Bonawitz et al., \citeyear{47bonawitz2019towards})(in Figure \ref{fig:Framework}), which has improved the design of the system architecture. In addition, the combination of blockchain and FL constitutes a BlockFL architecture. It enables the comparison of the performance of different terminals (Kim et al., \citeyear{48kim2019blockchained}). The following factors must be considered in the process of realizing these applications:
\begin{itemize}
\item \textbf {Privacy.} One of FL's core goals is to protect the private information of users. Recent research has shown that some malicious participants or FL servers may be present in the FL process, potentially resulting in privacy and security issues and generating corrupted global models. Malicious users can infer sensitive information, such as gender, occupation, and location, based on the sharing models of other participating users. Researchers use the FaceScrub dataset to train a binary gender classifier. In this process, by checking the shared model, the accuracy of inferring whether a participant's input is included in the dataset is found to be as high as 90\% (Melis
et al., \citeyear{49melis2019exploiting}).
\item \textbf {Security.} During the FL training process, the participating users train the learning model locally and share the training parameters with other participants which can improve forecast accuracy. However, they are often vulnerable to various attacks. For example, data and models go missing or become corrupted. In this attack mode, malicious users may send incorrect parameters or corrupt models. Thus, the global model will update incorrectly, and the entire learning system will be damaged. Simultaneously, the loopholes in the FL protocol may cause the risk of data privacy being destroyed. Lyu et al. analyzed and investigated the threat model and attack method under this behavior (Lyu et al., \citeyear{2020Threats}). This provides a guarantee for the safe application of data.
\end{itemize}

\textcolor{blue}{\subsection{Data application scenarios under IoT}}

\noindent At present, some scholars have proposed a novel FL framework for efficient communication and privacy protection which improves the performance of the IoT. Subsequently, it stabilized the dynamics data flow through Transmission Control Protocol and Cubic Curve Binary Increase Congestion (TCP CUBIC) flow on the Wi-Fi network. Finally, it obtained a good training model (Pokhrel \& Choi, \citeyear{50pokhrel2020improving}). The creation of a joint cloud video recommendation framework based on deep learning (DL) for mobile IoT meets the needs of users for applications. Simultaneously, it uses quantitative methods to reduce the uplink communication cost and network bandwidth (Duan et al., \citeyear{51duan2019jointrec}). In addition, FL enables resource-constrained edge computing devices to learn shared predictive models (Y. Zhao et al., \citeyear{31zhao2018federated}).

\textcolor{blue}{\subsection{Blockchain federated learning (Block FL)}}

\noindent The development of blockchain technology provides a new development direction for IoT. BlockFL architecture completes well the update of the local learning model. It uses a consensus mechanism and performs good performance data analyses (Kim et al., \citeyear{21kim2019blockchained}). In the industrial IoT, some researchers have designed a secure data sharing architecture authorized by blockchain. This process maintains data privacy effectively through a shared data model. Compared with real-world datasets, it has good accuracy, high efficiency and safety (Lu et al., \citeyear{53lu2019blockchain}).The existing FL method is based on the semi-honest assumption of the client to realize secure multi-party computation which is vulnerable to attacks from malicious clients. Awan et al. (\citeyear{2019Poster}) proposed a blockchain-based privacy-preserving federated learning (BC-based PPFL) framework. The framework is based on the immutability and decentralization of the blockchain. It implements safe updating of the local model and reliable data sources.

\textcolor{blue}{\subsection{Federated learning for edge computing}}

\noindent At present, FL has been combined with edge computing and achieves a good practical application. The use of edge and terminal computing can meet the requirements of cloud capacity and equipment at the edge of the network. Under this condition, FL has realized the application of a 4G/5G-based interconnected vehicle edge computing platform. This model completes the edge collaborative learning of real datasets collected by large electric vehicle (EV) companies. This method has the advantages of driver personalization, asynchronous execution and security protection. In addition, the personalized FL of the application of intelligent IoT can alleviate the negative impact of heterogeneity from different perspectives (Wu
et al., \citeyear{54wu2020personalized}). Simultaneously, a framework design based on FL can utilize limited bandwidth resources. We need to combine DL techniques and FL frameworks with mobile edge systems simultaneously  (Mcmahan
et al., \citeyear{57_2016Communication}). This can accelerate the application of mobile edge computing.

The existing implementation mode of FL allows computing nodes to synchronize only the local training model in distributed training. This leads to an FL architecture that relies on highly concentrated types and a large server bandwidth. However, the network capacity distribution between nodes is highly uniform, and is smaller than that of a data center. In (Jiang et al., \citeyear{55jiang2020bacombo}), the author proposes that the bandwidth between nodes can be used to speed up communication. First, it performs staff selection through segmented gossip aggregation and bandwidth awareness of the network. Second, it makes full use of the bandwidth between the nodes and between workers. This ultimately increases the convergence speed and reduces the number of communication rounds that are involved. Nowadays, the general FL system uses a central parameter server to coordinate a large federation of the participating workers. Workers use their own data sets to train local models. It periodically updates the parameters to the server for synchronization. The model updates of all nodes in the system also be sent to all other nodes. But this process will consume a lot of bandwidth resources and increase costs. Therefore, they used the model split-level synchronization mechanism. First, they divide a model into a set of segment subsets containing the same number of model parameters that do not overlap with each other. Second, the workers aggregate the partial divisions with the corresponding divisions of \textit{k} other workers. Then, it performs a segmentation level update. Third, they divided the other workers, which maximizes the bandwidth capacity between workers. It shares the communication cost and further accelerates the convergence speed.

In (X. Wang et al., \citeyear{56wang2019edge} ), the authors proposed to combine deep reinforcement learning techniques and FL frameworks with mobile edge systems. This can optimize mobile edge computing (MEC). In this process, the In-Edge AI framework was designed. It can intelligently use the cooperation between the device and the edge node to exchange learning parameters. Finally, it achieves dynamic system-level optimization and application-level enhancement. The key to solving this problem is that computing offloading requires wireless data transmission. The optimization of the entire communication and computing integration system jointly allocates the communication resources and computing resources of edge nodes. At the same time, it also uses floating and edge cache calculations between MEC systems. In addition, FL (B. Liu, Wang, \& Liu, \citeyear{liu2019lifelong}) has also been introduced as a framework for training agents in a distributed manner. The effects of this method are as follows: 1) reduces the amount of data that should be used; 2) responds to the mobile communication environment and cellular network conditions; 3) interacts with the actual cellular network, heterogeneous user equipment (UE) adapts well, and 4) protects personal data privacy.
\textcolor{blue}{\subsection{Challenges and problems}}

\noindent In Table \ref{table:method1}, the current researches provide certain solutions for the development of practical applications under the framework of IoT. But at the same time, there are many problems in the realization of FL, such as computing power, heterogeneity, security, and resource integration (X. Wang et al., \citeyear{56wang2019edge} ; Li, Sahu, Tal walkar, \& Smith, \citeyear{95li2020federated} ; Lim, Luong, et al., \citeyear{97lim2020federated} ;Khan et al., \citeyear{Khan2020FederatedLF}). They have an adverse effect on the development of IoT. So we put forward possible solutions to solve these problems. The specific plan is as follows:

\begin{itemize}
\item \textbf {Sparsification of FL:} The influence of factors such as wireless resource limitations and noisy datasets often affects the convergence of FL and the training of local models. We can construct a gradient-based sparsity scheme with integrating the available communication resources. Simultaneously, we clean up the data set and select equipment with sufficient computing power for training.
\item \textbf {Heterogeneous clustering of FL:} At present, a large number of equipment data sets have certain statistical heterogeneity. It greatly reduces the convergence performance of FL. We can choose terminal devices with a certain degree of trust to cluster in a group of datasets.
\item \textbf {Security of FL:} Malicious terminal devices may often be present during the training process. These wrong local learning model parameters will affect the accuracy and convergence time. We can try to use the blockchain to verify the update of the terminal equipment.
\item \textbf {Resource allocation of FL:} Terminal equipment will interfere with cellular users and occupy uplink communication resources in the process of FL. We can try to establish a resource allocation mechanism based on game theory. This one-to-many matching theory can effectively integrate resources. It links the resource block with other terminal devices that allocate resources.
\end{itemize}
\begin{table}[htpb]
\centering
\caption{Category, Key contributions, and Framework in applications of FL in IoT}
\label{table:method1}
\resizebox{\textwidth}{!}{%
\begin{tabular}{|c|c|l|c|}
\hline
Category &
  Reference &
  \multicolumn{1}{c|}{Key contributions} &
  Framework \\ \hline
\multirow{4}{*}{\begin{tabular}[c]{@{}c@{}}Mobile terminal\\ devices\end{tabular}} &
  (Y. Zhao et al., \citeyear{31zhao2018federated}) &
  \begin{tabular}[c]{@{}l@{}}
   1.Focused on the statistical challenges of FL  when the local data is non-IID.\\
   2.Calculated the earthmover's distance (EMD) of each device distribution \\ to quantify the weight difference.\\
   3.Created a small portion of data that is globally shared between edge \\ devices to improve  training on non-IID data.
  \end{tabular} &
  FL \\ \cline{2-4} 
 &
  (Bonawitz et al., \citeyear{47bonawitz2019towards}) &
  \begin{tabular}[c]{@{}l@{}}
  1.Provided a scalable production system for FL in the field of mobile devices\\ based on TensorFlow.\\
  2.Brought advanced designs for FL of mobile devices, such as: On-device \\item ranking, Content suggestions for on-device keyboards, Next word prediction.\end{tabular} &
  FL \\ \cline{2-4} 
 &
  (Duan et al., \citeyear{51duan2019jointrec}) &
  \begin{tabular}[c]{@{}l@{}}
  1.Proposed the JointRec federated cloud video recommendation framework.\\ 2.Reduced the uplink communication cost and network bandwidth.\end{tabular} &
  JointRec \\ \cline{2-4} 
 &
  (Mcmahan et al., \citeyear{57_2016Communication}) &
  \begin{tabular}[c]{@{}l@{}}1.Proposed a deep network FL based  on iterative model averaging.\\ 
  2.Compared with synchronous stochastic gradient  descent, the number of\\ communication rounds required  was greatly reduced.\end{tabular} &
  FL \\ \hline
\multirow{2}{*}{Blockchain} &
  (Kim et al., \citeyear{48kim2019blockchained}) &
  \begin{tabular}[c]{@{}l@{}}1.Used the blockchain to propose a BlockFL framework.\\ 
  2.Analyzed the end-to-end delay model of BlockFL.\\ 
  3.Described the optimal block generation rate from communication, calculation\\ and consensus delay.\end{tabular} &
  BlockFL \\ \cline{2-4} 
 &
  (Lu et al., \citeyear{53lu2019blockchain}) &
  \begin{tabular}[c]{@{}l@{}}1.Proposed a secure data sharing architecture  authorized by blockchain.\\ 
  2.Converted data sharing problems into ML problems by merging\\ and preserving privacy-preserving FL.\\ 
  3.Used consensus calculations for training.\end{tabular} &
  BlockFL \\ \hline
IoV &
  (Pokhrel \& Choi, \citeyear{50pokhrel2020improving}) &
  \begin{tabular}[c]{@{}l@{}}1.Proposed a efficient FL framework to improve the performance of the Internet\\ of Vehicles (IoV).\\
  2.Considered the TCP CUBIC flow on the WiFi network for verification and\\ stabilize its data flow dynamics.\end{tabular} &
  FL \\ \hline
\multirow{2}{*}{\begin{tabular}[c]{@{}c@{}}Edge \\ \\ Computing\end{tabular}} &
  (Wu et al., \citeyear{54wu2020personalized}) &
  \begin{tabular}[c]{@{}l@{}}1.Promoted a personalized FL framework for smart IoT applications in the\\ cloud edge architecture.\\ 
  2.Reduced the negative impact of heterogeneity in the training process.\\ 3.Realized fast processing and low latency through edge computing.\end{tabular} &
  FL \\ \cline{2-4} 
 &
  (X. Wang et al., \citeyear{56wang2019edge}) &
  \begin{tabular}[c]{@{}l@{}}1.The “In-Edge AI” framework was designed to use the collaboration between\\ devices and edge nodes to exchange learning parameters.\\ 2.Realized dynamic system-level optimization and  application enhancement.\\ 3.Reduced unnecessary system communication load.\end{tabular} &
  In-Edge AI \\ \hline
\begin{tabular}[c]{@{}c@{}}IoT \\ communication\end{tabular} &
  (Jiang et al., \citeyear{55jiang2020bacombo}) &
  \begin{tabular}[c]{@{}l@{}}1.Proposed Bandwidth Aware Combination (BACombo) to solve the network\\ capacity between computing nodes.\\ 
  2.This mechanism made full use of the node-to-node bandwidth to speed up\\ communication time.\\ 3.Reduced transmission delay and reduced training time while ensuring accuracy.\end{tabular} &
  BACombo \\ \hline
\begin{tabular}[c]{@{}c@{}}Cloud Robot\\  System\end{tabular} &
  (B. Liu, Wang, \& Liu, \citeyear{liu2019lifelong}) &
  \begin{tabular}[c]{@{}l@{}}1.Proposed a learning architecture for navigating in cloud robot systems: Lifelong \\Joint Reinforcement Learning (LFRLA).\\ 2.Improved the efficiency of reinforcement learning for robot navigation.\end{tabular} &
  LFRLA \\ \hline
\end{tabular}%
}
\end{table}

\textcolor{blue}{\section{Application of federated learning to intelligent transportation system in smart cities}}

\begin{figure}[thpb]
	\centering
	\includegraphics[width=0.9\linewidth]{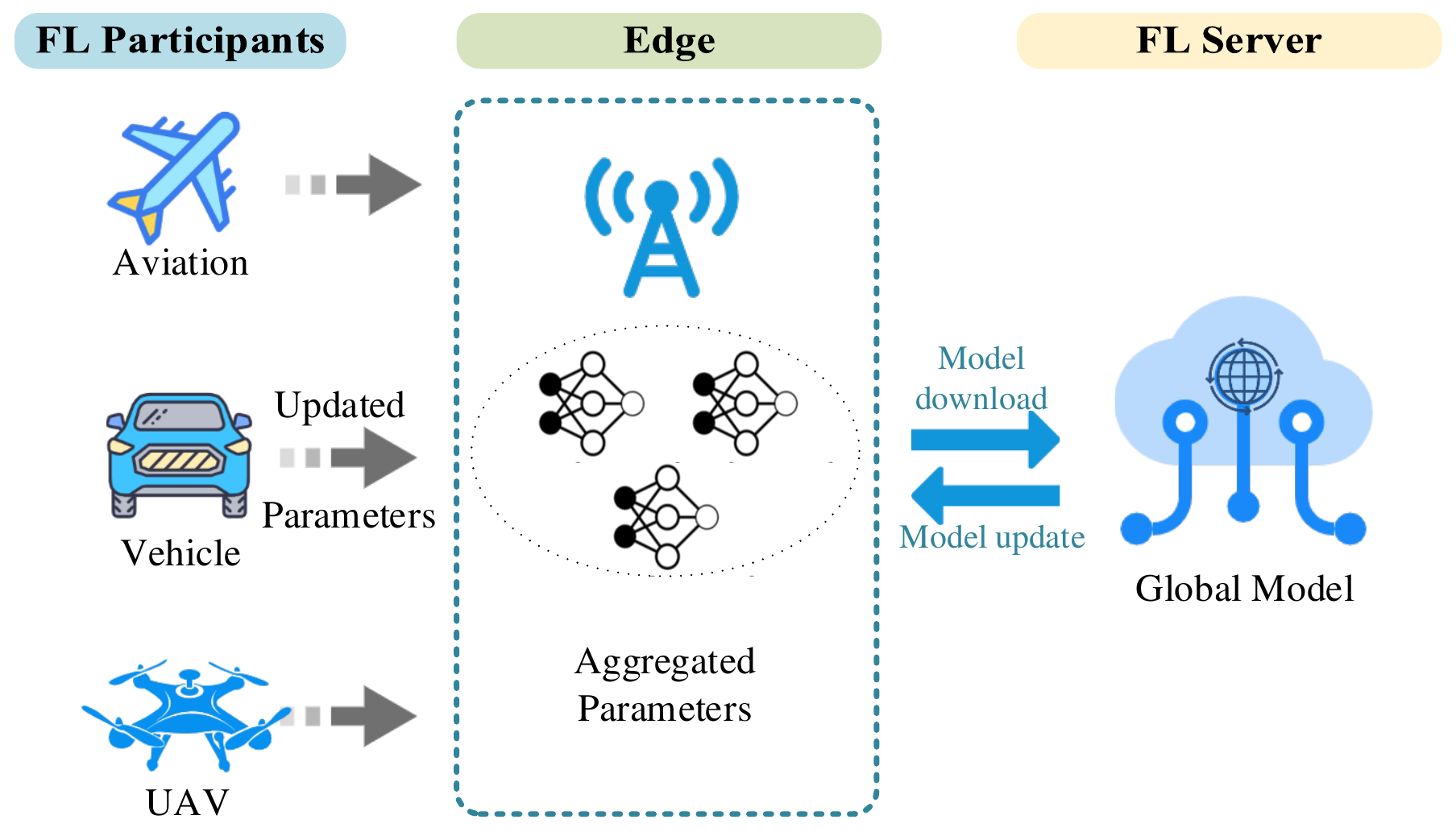}
	\caption{Application of federated learning in transportation systems.}
	\label{fig:transportation}
\end{figure}

\noindent Transportation is an integral part of a smart city. As shown in Figure \ref{fig:transportation}, we can solve various problems in transportation systems through FL, such as communication delays, calculation processing, and data privacy.

\textcolor{blue}{\subsection{Combination of federated learning and vehicle system}}
\textcolor{blue}{\subsubsection{Vehicle communication}}

\noindent Emerging vehicle applications highly depend on vehicle-to-vehicle (V2V) communication. Therefore, it is necessary to consider URLLC in vehicle networks when developing intelligent transportation systems (Ashraf et al., \citeyear{58ashraf2017towards}; Pokhrel \& Choi, \citeyear{50pokhrel2020improving}).
Samarakoon et al. (\citeyear{62samarakoon2018federated}) described the problem of joint power control and resource allocation in vehicle-mounted communication networks as a net-range power minimization problem constrained by URLLC. Vehicular users (VUEs) estimate the tail distribution locally with roadside units (RSU) (Xu et al., \citeyear{xu2020blockchain}). The constraint of URLLC is characterized by extremum theory and is modeled as a tail distribution of the network scope queue length over a predefined threshold. It can effectively reduce delays and enhance reliability (Samarakoon et al., \citeyear{22samarakoon2019distributed}). However, cloud-based learning methods are relatively slow. The author (Pokhrel \&
Choi, \citeyear{50pokhrel2020improving}) proposed a systematic IoT network design approach, which accelerates the learning process of data transmission protocols (e.g., TCP) that convert vehicles into mobile data centers.

\textcolor{blue}{\subsubsection {Electric vehicle}}

\noindent In the future, the large-scale use of EVs is inevitable. It will generate enormous energy demand. Therefore, maintaining effective energy demand forecasting services for charging station (CS) providers is an urgent problem. Due to privacy protection, toll CSs and vehicle companies cannot share data. X. Wang et al. ( \citeyear{wang2021charging}) uses the features of the data on both sides and the cross features between the two to build the model through encrypted entity alignment, secure FL, and prediction. A model with cross-features is also introduced, and Area Under the Curve (AUC) is improved. Finally, the result of relatively centralized learning is almost lossless. Saputra et al. (\citeyear{saputra2020federated}) introduced a CSs-based decentralized federated energy learning (DFEL) framework to learn local datasets through CSs to predict energy requirements accurately and reduce communication cost markedly. 

\textcolor{blue}{\subsubsection {Autonomous vehicle}}

\noindent For autonomous vehicles(B. Liu, Wang, Liu, \& Xu, \citeyear{liu2019federated}), we should maintain their best ML models and the ability to make intelligent decisions. Pokhrel \& Choi (\citeyear{pokhrel2020federated}) proposed an autonomous blockchain-based federated learning (BFL) design, which uses the consensus mechanism of the blockchain to enable on-Vehicle Machine Learning (oVML). Their reward method developed a mathematical framework with controllable network and BFL parameters to investigate system-level performance effects. It is proved that the idea of adjusting the block arrival rate is provably online. In conjunction with a contract-theoretic incentive mechanism designed, Zeng et al. (\citeyear{zeng2021federated}) proposed an FL framework for collaborative learning and optimization of its autonomous controller design under the conditions of wireless link uncertainty and environmental dynamics.

\textcolor{blue}{\subsection{Combination of federated learning and aviation system}}
\textcolor{blue}{\subsubsection{Aircraft}}

\noindent Owing to the large amount of data generated by the aviation system and the lack of computing resources, it cannot perform the fault prediction of aircraft. Moreover, the deployment of additional airborne resources is very complicated and expensive. Therefore, in (Aussel et al., \citeyear{66aussel2020combining}), the authors proposed a method of using an active online decision tree based on confidence as the basic model of client learning (De Rosa \& Cesa Bianchi, \citeyear{67de2017confidence}). They classified standard samples with minimum computing power and established a mechanism for transmission and identification of uncertain data under the communication budget.

\textcolor{blue}{\subsubsection{Unmanned aerial vehicle(UAV)}}

\noindent Due to limitations in computing and power resources, traditional centralized DL has led to lower network bandwidth and UAVs' energy efficiency. Brik et al. (\citeyear{brik2020federated}) discussed how to use federated deep learning (FDL) to deal with target challenges in wireless networks supported by UAVs, as well as the critical technical challenges open problems of FDL-based methods and future research directions. Lim, Huang, et al. (\citeyear{lim2020towards}) proposed FL-based sensing and collaborative learning solutions through contract matching incentive design so that the lowest cost UAVs can be matched to each partition. Besides, We have noticed that there are few studies on the battery problems of FL for UAVs. Tang et al. (\citeyear{tang2021battery}) adjusted its operating CPU frequency to extend battery life and promptly quit FL. With the strategy based on Deep Deterministic Strategy Gradient (DDPG), they combined delay and energy consumption linearly to estimate the system cost. All devices can complete all FL tasks with limited batteries while reducing system costs significantly.

\textcolor{blue}{\subsection{Challenges and problems}}

\noindent With the development of science, the characteristics of the transportation system are constantly changing. After investigating the application in the FL transportation system, we list the innovations and contributions of FL in transportation systems in Table \ref{table:transportation}, the challenges and problems are described as below:

\textcolor{blue}{\subsubsection{Communication and calculation costs}}

\noindent Since the mobility of the equipment in the transportation system, resources in communication and computing are limited (Aussel et al., \citeyear{66aussel2020combining}; De Rosa \& Cesa-Bianchi, \citeyear{67de2017confidence}). Therefore, the challenge is how to reduce the communication and computing overhead of FL, successfully apply the data to other learning scenarios (Saputra et al., \citeyear{saputra2020federated}; Brik et al., \citeyear{brik2020federated}), improve the design models' accuracy and efficiency (Zeng et
al., \citeyear{zeng2021federated}), and avoid affecting the performance of the framework.

\textcolor{blue}{\subsubsection{Privacy protection}}

\noindent With the complexity of the calculation method violated the privacy of users constantly changing, when the miners tried to verify the local models (Pokhrel \& Choi, \citeyear{pokhrel2020federated}), the risk of privacy leakage will greatly increase. It is an option to provide a dynamic and scalable method through risk analysis to protect user privacy from the impact of the construction of large data sets (X. Wang et al., \citeyear{wang2021charging}).

\textcolor{blue}{\subsubsection{Energy issues}}

\noindent With the increasing depletion of energy, clean energy has become the first choice. In EVs and UAVs, and other transportation equipment that uses electric energy, we need to study the balance between energy saving and FL performance (Tang et al., \citeyear{tang2021battery}; Saputra et al., \citeyear{saputra2020federated}).

\begin{table}[htpb]
  \centering
  \caption{Category, Key contributions, and Framework in applications of FL in the field of intelligent transportation system in smart city.}
  \label{table:transportation}
  \resizebox{\textwidth}{!}{
\begin{tabular}{|c|c|l|c|}
\hline
Category &
  Reference &
  \multicolumn{1}{c|}{Key   contributions} &
  Framework \\ \hline
\multirow{4}{*}{\begin{tabular}[c]{@{}c@{}}Resource \\ allocation\end{tabular}} &
   (Samarakoon et al., \citeyear{22samarakoon2019distributed}) &
   \begin{tabular}[c]{@{}l@{}}1.Proposed a distributed, FL-based joint transmit power\\ and resource allocation framework. \\ 2.Proposed an asynchronous FL algorithm for maximum\\ likelihood estimation.\end{tabular} &
  FL \\ \cline{2-4} 
 &
  (Samarakoon et al., \citeyear{62samarakoon2018federated}) &
  \begin{tabular}[c]{@{}l@{}}1.Proposed a clean-slate design of resource allocation. \\ 2.Proposed a roadside unit-assisted approach. \\ 3.Designed a new virtual zone formation approach.\end{tabular} &
  FL \\ \cline{2-4} 
 &
  (Aussel et al., \citeyear{66aussel2020combining}) &
  \begin{tabular}[c]{@{}l@{}}1.Realized real-time distributed learning of aviation system. \end{tabular} &
  FL \\ \cline{2-4} 
 &
  (Ng et al., \citeyear{70ng2020joint}) &
  \begin{tabular}[c]{@{}l@{}}1.Improved communication efficiency between model\\ owners and workers. \end{tabular} &
  FL \\ \hline
\multirow{3}{*}{\begin{tabular}[c]{@{}c@{}}Privacy \\ protection\end{tabular}} &
  (Pokhrel \& Choi, \citeyear{pokhrel2020federated}) &
  \begin{tabular}[c]{@{}l@{}}1.Developed a comprehensive mathematical analysis of\\system dynamics for end-to-end delay analysis. \end{tabular} &
  BlockFL \\ \cline{2-4} 
 &
  (X. Wang et al., \citeyear{wang2021charging}) &
  \begin{tabular}[c]{@{}l@{}}1.Realized EV charging point recommendation by designing\\ the cross-platform FL architecture. \\ 2.An encrypted entity alignment method is proposed for\\different IDs from different platforms\end{tabular} &
  FL \\ \cline{2-4} \hline
\multirow{3}{*}{\begin{tabular}[c]{@{}c@{}}Energy\\Issue \end{tabular}} &
  (Saputra et al., \citeyear{saputra2020federated}) &
  \begin{tabular}[c]{@{}l@{}}1.Reduce communication overhead and \\ increase learning speed. \\ 2.Developed an iterative energy contract algorithm.\end{tabular} &
  Energy-efficient \\ \cline{2-4} 
 &
  (Tang et al., \citeyear{tang2021battery}) &
  \begin{tabular}[c]{@{}l@{}}1.Proposed a resource allocation strategy for UAVs based\\ on edge computing.\\  2.Proposed a resource allocation strategy based on deep\\reinforcement learning. \end{tabular} &
  FL \\ \cline{2-4} \hline
\end{tabular}}
\end{table}

\textcolor{blue}{\section{Federated learning in the financial field of smart cities}}

\noindent The financial field includes banking, insurance, trust, securities, and leasing. However, in recent years, there have been criminal activities in these sectors. Some financial crimes can involve up to several hundreds of millions of dollars, such as the mortgage crisis. These activities have led to crisis scenarios for families and society as a whole. While the financial industry expends a lot of resources annually to combat financial fraud, it is not very effective.

\begin{figure}[thpb]
	\centering
	\includegraphics[width=0.7\linewidth]{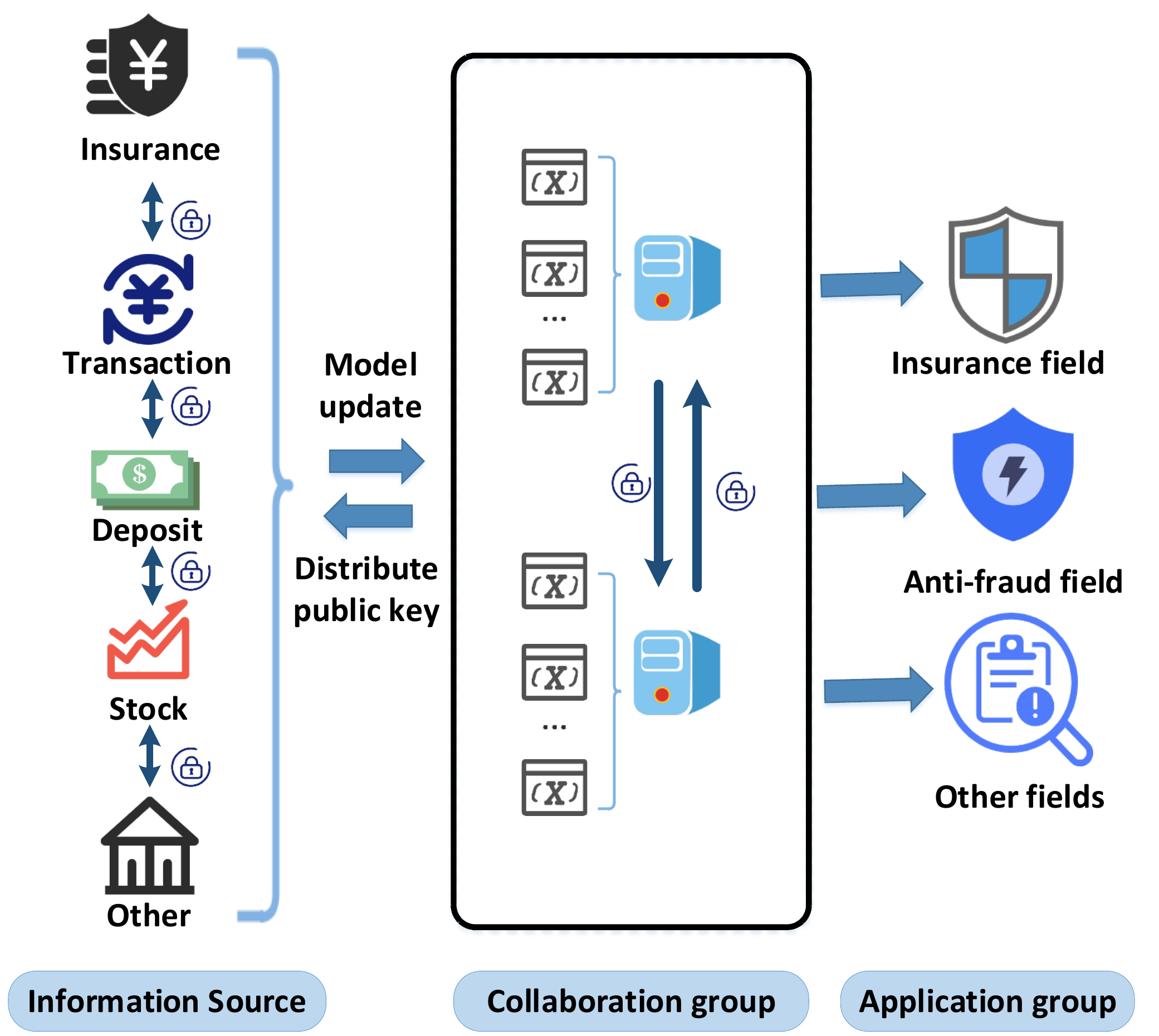}
	\caption{The processing framework of FL in the financial field. }
	\label{fig:financial}
\end{figure}

In recent years, it has become necessary to use FL to reduce losses to banks and consumers. In DL, the sample size must be sufficient to enable the training of a better model. A single bank cannot provide sufficient information on a person's consumption and credit cards, and it is also difficult for a bank to detect fraud. The concept of FL provides the financial industry with a new approach to training models using DL whereby the owner of each set of data can collaborate on the model without sharing the customer's private information. The financial industry faces several challenges when reviewing user qualifications and screening quality customers. The combination of FL and finance can effectively address this challenge while also protecting from disclosure customers' private information. As shown in Figure \ref{fig:financial}, the collaborative group sends the key to the data source. Then, the exchanged data are encrypted to ensure data security. Finally, the learning model is updated in real time to obtain the model output for different application scenarios. FL can be very effective in these areas. For example, banks can use cameras to identify suspicious transactions and prevent malicious multi-party lending (Y. Liu, Huang,
et al., \citeyear{72liu2020fedvision}; Q. Yang et al., \citeyear{73yang2019federated}). 

\textcolor{blue}{\subsection{Federated learning in the field of financial fraud}}

\noindent With the advent of the digital age, there have been many transnational financial crimes. Common sub-categories of financial crime are financial theft, fraudulent loans, and money laundering. Credit card fraud causes large losses to both banks and consumers. Chuan et al. (\citeyear{75ma2020safeguarding}) also provided satisfactory answers to some questions (model aggregation, data poisoning, scaling up issues) . In addition to improving FL, some studies have combined FL with other algorithms. 

In recent years, there has been an increasing interest in privacy issues (X. Ma et al.,
\citeyear{ma2020secure}). Wealthy individual are concerned about whether their private financial information can be effectively protected. Owing to the development of ML, an increasing amount of bank user data is analyzed and trained in relevant bank marketing models. However, protecting the private information of financial users is an important research direction. Feng Yan et al. (\citeyear{77feng2020practical}) proposed a bilateral privacy-protected FL scheme that also protects the iterative parameters during the training process. This scheme further protects model parameters from being acquired by external attackers on the basis of traditional FL, considering only the privacy of the client. 

\textcolor{blue}{\subsection{Federated learning in the field of insurance}}

\noindent Building a data service platform for the insurance industry requires the integration of financial, medical, and other data from multiple parties. If an insurance company wants to improve its risk management capability and business development level, it needs to consider the impact of multi-party data. The effective use of data without infringing on personal privacy is also an important issue in the insurance industry. \'Smietanka  et al. (\citeyear{smietanka2020algorithms}) proposed that the key technologies that promote the insurance industry's reform include FL and computable insurance contracts. Yuan et al. (\citeyear{liang2020isolated}) proposed a configurable FL benchmark suite, FLBench. This kit can simulate various isolated data islands according to specific research requirements, and covers areas such as insurance and securities.

When different insurance companies and multi-party data providers implement FL, the quantification of participant contributions is also a realistic problem. Wang et al. ( \citeyear{wang2019measure}) proposed that the service model can employ related models in order to use data models to integrate information to obtain better feedback. Yan et al. (\citeyear{liu2020real}) proposed an online evaluation method that is more sensitive to the quality and quantity of data, and compared it with the results obtained by the Shapley value in game theory. Table \ref{financial} shows the application and development of FL in the financial field.

\begin{table}[htpb]
  \centering
  \caption{Category, Key contributions, and Framework in applications of FL in the financial field.}
  \label{financial}
  \resizebox{\textwidth}{!}{
\begin{tabular}{|c|c|l|c|}
\hline
Category &
  Reference &
  \multicolumn{1}{c|}{Key contributions} &
  Framework \\ \hline
\multirow{4}{*}{\begin{tabular}[c]{@{}c@{}}Privacy \\ protection\end{tabular}} &
  (W. Yang et al., \citeyear{74yang2019ffd}) &
  \begin{tabular}[c]{@{}l@{}}1. Detection framework FFD.\\ 2. Using real data for testing, the average test AUC\\  of FDS based on joint learning reached 95.5\%.\end{tabular} &
  FL \\ \cline{2-4} 
 &
  (X. Ma et al., \citeyear{ma2020secure}) &
  \begin{tabular}[c]{@{}l@{}}1. A new framework for secure multi-party learning. \\ 2. A specific scheme is constructed by merging \\ aggregated signature and proxy re-encryption \\ technology.\end{tabular} &
  SML \\ \cline{2-4} 
 &
  (Y. Liu, Zhang, \& Wang, \citeyear{78liu2020asymmetrically}) &
  \begin{tabular}[c]{@{}l@{}}1. Asymmetrical federated model training.\\ 2. Innovatively proposed a genuine with dummy \\ approach to achieving asymmetrical federated \\ model training.\end{tabular} &
  FL \\ \cline{2-4} 
 &
  (Han et al., \citeyear{han2021verifiable}) &
  \begin{tabular}[c]{@{}l@{}}1. The key exchange technology.\\ 2. Double shielding protocol is used to ensure that\\ users' privacy is not leaked.\end{tabular} &
  FL \\ \hline
\multirow{3}{*}{\begin{tabular}[c]{@{}c@{}}Model fusion\\  method\end{tabular}} &
  (C. Ma et al., \citeyear{75ma2020safeguarding}) &
  1. An intelligent aggregation method. &
  FL \\ \cline{2-4} 
 &
  (Gu et al., \citeyear{gu2020privacy}) &
  \begin{tabular}[c]{@{}l@{}}1. FL algorithms for vertically partitioned data.\end{tabular} &
  FL \\ \cline{2-4} 
 &
  (Yu et al., \citeyear{yu2020fed+}) &
  \begin{tabular}[c]{@{}l@{}}1. A new federated learning framework, Fed+, was\\introduced.\\ 2. Better handle the statistical heterogeneity inherit\\in the federated environmen\end{tabular} &
  Fed+ \\ \hline
\multirow{3}{*}{\begin{tabular}[c]{@{}c@{}}Financial \\ applications\end{tabular}} &
  (Suzumura et al., \citeyear{76suzumura2019towards}) &
  \begin{tabular}[c]{@{}l@{}}1. A method for sharing key informationbetween \\ institutions by using theFederated Graph Learning \\ platform isproposed.\\ 2. The constructed model has aperformance that \\ is 20\% higher thanthe original model.\end{tabular} &
  FL \\ \cline{2-4} 
 &
  (Zheng et al., \citeyear{zheng2021asfgnn}) &
  \begin{tabular}[c]{@{}l@{}}1. The author proposed an AutomatedSeparated-\\ Federated Graph NeuralNetwork(ASFGNN) learning \\ example.\\ 2. Achieve better results in actual experiments.\end{tabular} &
  ASFGNN \\ \cline{2-4} 
 &
  (Long et al., \citeyear{long2020federated}) &
  \begin{tabular}[c]{@{}l@{}}1. Propose Open banking.\\ 2. For the statistical heterogeneity, this article has\\a deep discussion\end{tabular} &
  FL \\ \hline
\end{tabular}}
\end{table}

\textcolor{blue}{\subsection{Challenges and problems}}

\noindent Nowadays, financial field, which has a huge amount of structured data, provides a perfect foundation for the implementation of AI. However, due to the particularity of the financial industry, it has stricter requirements for data security and privacy. In addition to the general FL environment problems, this article believes that FL needs to be solved urgently in the financial field as follows.

Statistical challenges. The data distribution of different companies in the financial industry varies greatly. For example, the size of the company varies greatly depending on the population and age group. Most financial companies suffer from data confusion and efficiency issues. The quality of data collected from multiple sources is uneven, and there is no uniform data standard and scale. How to effectively count the similarities and differences of different data sources is still the key to FL in the financial field.

Incentive mechanism. Large financial companies have mastered more user information and small and medium-sized financial companies have more valuable information and are often unwilling to share data. If there is no design incentives, FL cannot develop well in the financial field. Therefore, how to establish effective incentive measures to attract high-quality financial data into the FL system is an urgent problem to be solved.

\textcolor{blue}{\section{Application of federated learning to the medical field in smart cities}}

\noindent With the rapid increase of COVID-19 worldwide, the burden on medical staff has gradually increased. The treatment of patients using effective methods is a major problem. Smart medicine is an area of future medical development, and this area is expected to benefit from the increased development of FL technology. In the past, owing to the independence of hospitals and the privacy of information, there was a lack of sufficient samples for ML. On the right side of Figure \ref{fig:medical}, FL can unite previously independent individual hospitals into a collective population, significantly increasing the sample size of model training.

\begin{figure}[thpb]
	\centering
	\label{fig:medical}
	\includegraphics[width=1\linewidth]{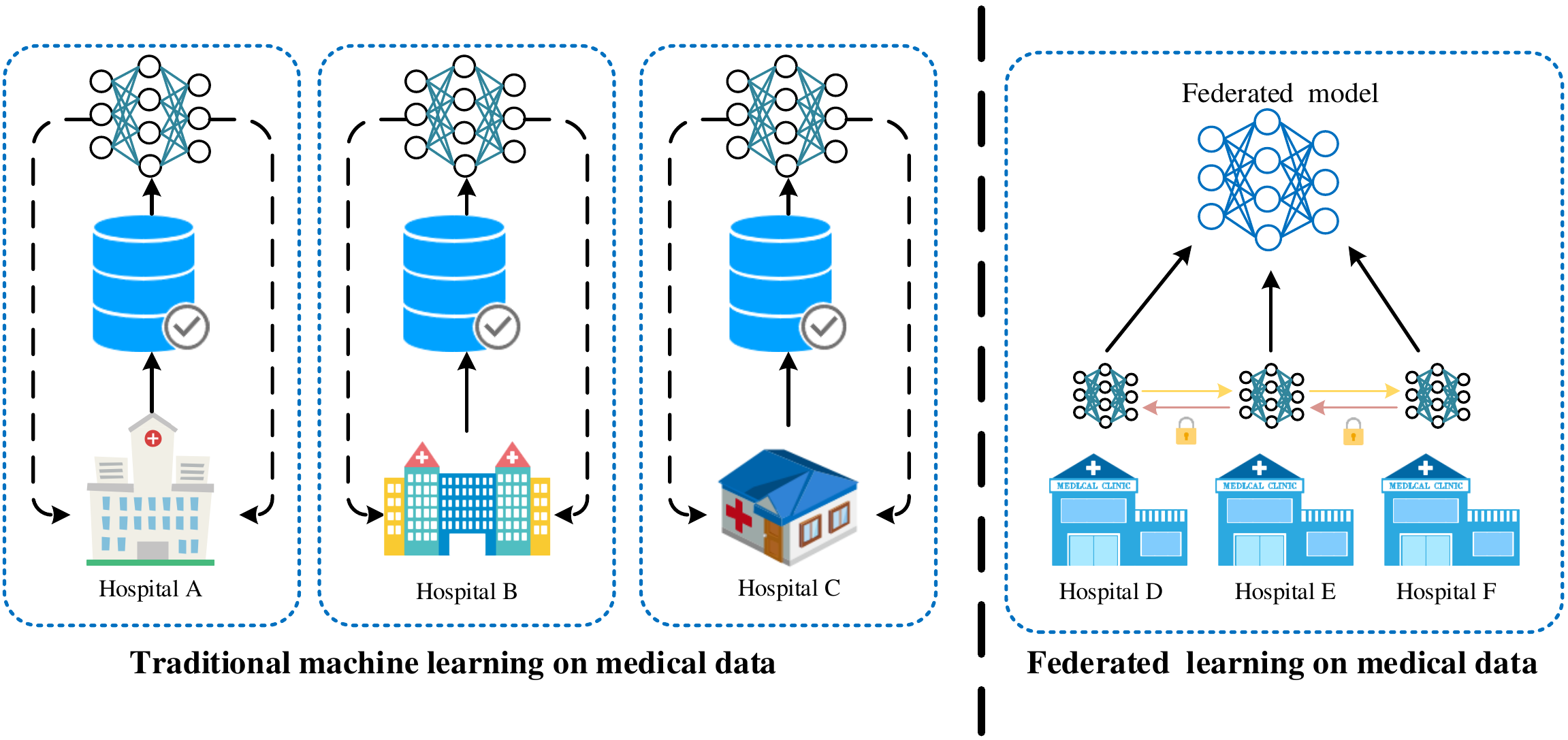}
	\caption{Comparison of training processes of traditional ML on medical data and FL on medical data.}
	\label{fig:medical}
\end{figure} 

\textcolor{blue}{\subsection{Combination of federated learning and medical privacy protection}}

\noindent With the Precision Medicine Initiative in the United States and the emergence of a large amount of personal health electronic information, patient data are usually protected in localized silos. However, there is an increasing desire to merge datasets from different medical systems. Because the constraints of establishing a calibration model locally may limit the degree of improvement, Huang (\citeyear{79huang2020preservation}) proposed a SMC method to establish a global isotonic regression calibration model. Fang et al.  (\citeyear{80fang2019redsync}) proposed a method for reducing the transmission bandwidth and protecting privacy in distributed learning. 

It is worth noting that Vepakomma  et al. (\citeyear{81vepakomma2019reducing}) further demonstrated the minimization of the distance correlation between the original data and intermediary representation. This reduces the leakage of sensitive raw data patterns during client communication, while maintaining the accuracy of the model. It reduces the leakage of the communication payload and original data present in the medical data. 
 Chamikara et al. (\citeyear{chamikara2020privacy}) proposed a privacy-protected FL framework for multi-site functional MRI analysis, and they studied the use of brain function connections to classify the communication speed and privacy protection of autism spectrum disorders and health control problems.

\textcolor{blue}{\subsection{Combination of federated learning and drug development}}

\noindent FL has revolutionized leading fields, including health care technology, and it has resulted in outstanding achievements in many fields, such as drug discovery. The proposed FL approach allows the pharmaceutical industry to use distributed data from different sources without leaking sensitive information on these data. This emerging decentralized machine-learning paradigm is expected to significantly improve the success of artificial intelligence drug discovery. Chen et al. (\citeyear{86chen2020fl}) verified the feasibility of applying horizontal federated learning (HFL). FL quantitative structure-activity relationship (FL-QSAR) under the HFL framework provides an effective way to break the barriers of pharmaceutical institutions in QSAR modeling. The solution promotes the development of collaboration and privacy-preserving drug discovery, and it can be extended to other privacy-related biomedical fields. Xiong et al. (\citeyear{85xiong2020facing}) demonstrated the application of FL in predicting drug-related properties. Meanwhile, they also emphasized its potential role in solving small data and biased data dilemmas in drug discovery.

\textcolor{blue}{\subsection{Federated learning combined with disease prediction}}

\noindent FL has been proven to be an effective way of helping the medical industry make decisions and predict diseases. FL can further expand the sample size and protect privacy. In the medical field, more accurate judgments are often required for disease prediction and decision-making. For example, in the detection of lung nodules, the lung nodes are often too small to be detected. If it is incorrectly assessed, it is likely to reduce the patient's chances of survival. FL helps to produce enhanced prediction results and can protect data privacy and security. Table \ref{medical} is for the application and innovation of FL in the medical field.

\begin{table}[htpb]
  \centering
   \caption{Category, Key contributions, and Framework in applications of FL in the medical field}
     \label{medical}
   \resizebox{\textwidth}{!}{
\begin{tabular}{|c|c|l|c|}
\hline
Category &
  Reference &
  \multicolumn{1}{c|}{Key   contributions} &
  Framework \\ \hline
\multirow{2}{*}{\begin{tabular}[c]{@{}c@{}}Medical\\ imaging\end{tabular}} &
  (B. Liu, Yan, Zhou, Yang, \& Zhang, \citeyear{liu2020experiments}) &
  \begin{tabular}[c]{@{}l@{}}1. It is based on the FL framework testing the performance of\\ five models.\\ 2. The performance of the ResNet18 model is the best in comparison.\end{tabular} &
  FL \\ \cline{2-4} 
 &
  (Z. Yan et al., \citeyear{yan2020variation}) &
  \begin{tabular}[c]{@{}l@{}}1. Proposed a more stable variation-aware FL (VAFL) framework.\\ 2. Classification of prostate cancer data is better than FL \\ framework.\end{tabular} &
  FL \\ \hline
\multirow{4}{*}{\begin{tabular}[c]{@{}c@{}}Medical\\ data\end{tabular}} &
  (Rajendran et al., \citeyear{rajendran2021cloud}) &
  \begin{tabular}[c]{@{}l@{}}1.The FL framework is tested on two data samples and performance\\ is measured.\\ 2. The FL method generally does not improve the performance of\\ logistic regression\end{tabular} &
  FL \\ \cline{2-4} 
 &
  (Vaid et al., \citeyear{vaid2021federated}) &
  \begin{tabular}[c]{@{}l@{}}1. the effectiveness of multiple models for mortality prediction\\ under the data of five hospitals was tested.\end{tabular} &
  FL \\ \cline{2-4} 
 &
  (Lee \& Shin, \citeyear{lee2020federated}) &
  \begin{tabular}[c]{@{}l@{}}1. FL shows reliable performance under unbalanced, skewed and \\ extreme data distribution.\end{tabular} &
  FL \\ \cline{2-4} 
 &
  (Ge et al., \citeyear{ge2020fedner}) &
  \begin{tabular}[c]{@{}l@{}}1. A privacy-preserving medical NER method based on federated \\ learning.\end{tabular} &
  FL \\ \hline
\multirow{2}{*}{\begin{tabular}[c]{@{}c@{}}Data\\ processing\end{tabular}} &
  (Zhang et al., \citeyear{zhang2020dynamic}) &
  \begin{tabular}[c]{@{}l@{}}1. A dynamic fusion method is proposed, and the model fusion is\\ arranged according to the training time of the participating \\ customers.\\ 2. The categories of medical diagnostic image data sets used for\\  COVID-19 detection are summarized.\end{tabular} &
  FL \\ \cline{2-4} 
 &
  (Sui et al., \citeyear{sui2020feded}) &
  \begin{tabular}[c]{@{}l@{}}1. The strategy based on knowledge extraction was used to overcome\\  the communication bottleneck in FL.\\ 2. The article produced satisfactory results on three different \\ medical data sets.\end{tabular} &
  FL \\ \hline
\end{tabular}}
\end{table}

\textcolor{blue}{\subsection{Challenges and problems}}

\noindent In the above, we compared FL's latest technology and direction in the medical field. However, there are still many challenges for the development of FL in the medical field. This article believes that the most urgent problems to be solved in implementing medical FL are as follows. 

Heterogeneity of medical data. FL can fuse medical data, but it is one of the most challenging problems to mix horizontal medical data from medical institutions in different regions and longitudinal medical data of the same patient in various hospitals.

Model accuracy and diversity. The development of AI has been mostly realized in a closed scene. For example, AI challenges professional Go players, and AI game play is implemented in a known environment. However, in the intelligentization of medical treatment, there are many unexpected symptoms.

Dirty data identification. In the medical model, there are also misdiagnosis by doctors or interference by malicious data. For example, Malekzadeh (\citeyear{malekzadeh2021dopamine}) and others have some methods to distinguish between benign models and malicious models, they have strong limitations. It can only process malicious data in a specific environment.

\textcolor{blue}{\section{Application of federated learning to the communication field in smart cities}}

\begin{figure}[thpb]
	\centering
	\includegraphics[width=0.9\linewidth]{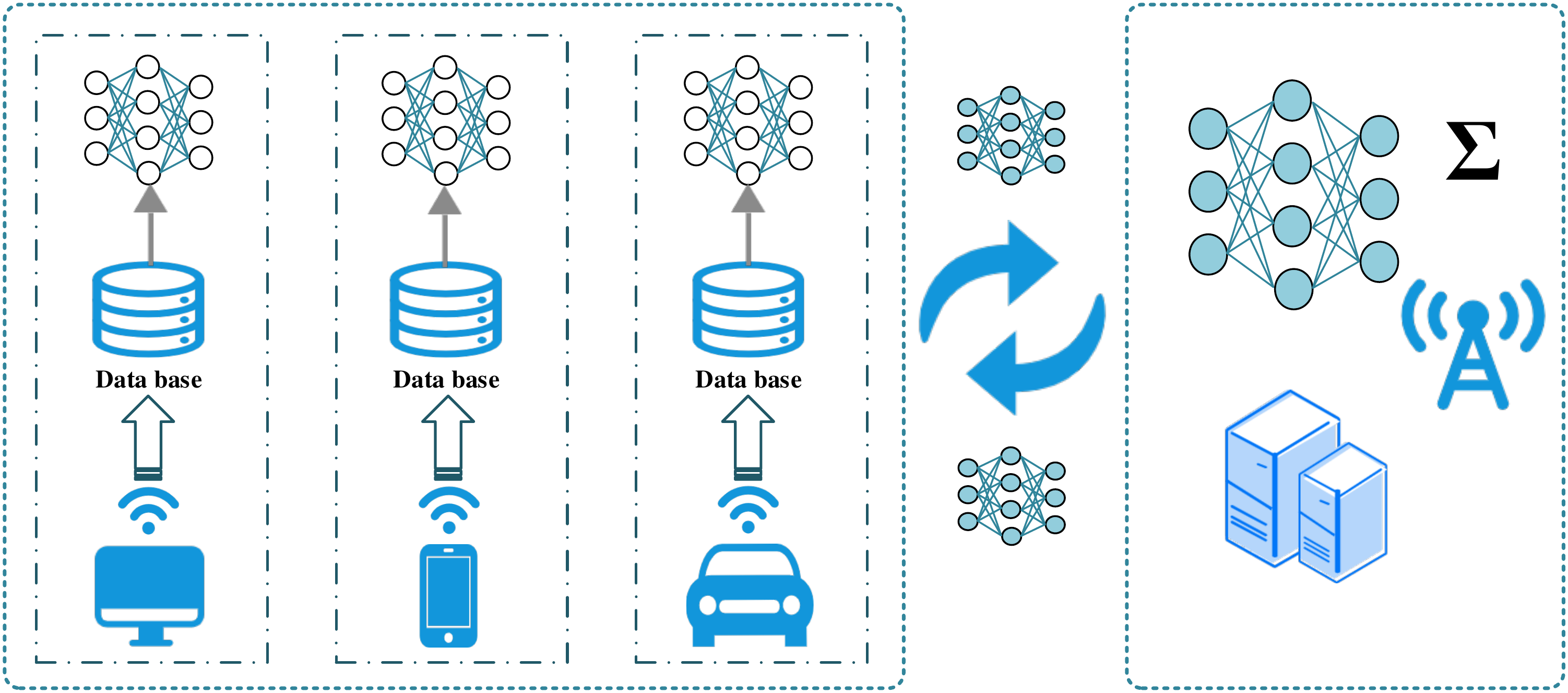}
	\caption{Network interaction diagram of federated learning in communication.}
	\label{fig:communication}
\end{figure}

\noindent FL allows distributed machines or users to cooperate in training machine-learning models with the help of parameter servers. It periodically updates the centralized server to protect user privacy. However, with the development of DL, massive datasets are needed to achieve better ML. The participants and servers require several rounds of communication to achieve the target accuracy. This process may result in the need for several million parameters and high communication cost (K. He et al., \citeyear{88he2016deep}). In addition, there are problems such as delays in IoT terminal equipment (Luping et al., \citeyear{89luping2019cmfl}), instability of communication links, and lack of data links and computing resources in aviation communications. The transportation system is inefficient and expensive. At present, FL has led to significant breakthroughs in the field of multiple-access channel communication(in Figure \ref{fig:communication}). 
One of the challenges faced by FL is the required communication overhead owing to its iterative nature and large model size. One new method to alleviate the bottleneck of FL communication is to allow the simultaneous display of user traffic on multiple access channels; this may improve the use of communication resources. Another is to explore the superposition characteristics of the wireless multiple-access channel to calculate the required function of the distributed local calculation update (i.e., the weighted average function).

\textcolor{blue}{\subsection{Federated learning solution for multiple access channel problem}}

\noindent Previous work relieved the communication bottleneck by compressing the gradient before transmission. Two commonly used methods are (A) quantization and (B) sparse gradient quantization. It follows the lossy compression idea of using a small number of bits to describe the gradient. These low-precision gradients are transmitted back to the parameter server (PS). However, these independent compression techniques have not been adjusted to the underlying communication channel exchanged between the user and the parameter server, and channel resources may not be fully utilized. Another study of FL through wireless channels is a more general multiple-access channel. The stacked nature of the wireless channel allows gradients to be clustered together in the air, and enables more effective training. These methods can be roughly classified as digital or analog solutions depending on how the gradient is transmitted through the channel. In the simulation scheme, the local gradient is scaled and transmitted directly through the wireless channel. In the digital scheme, the slave users are decoded separately, but transmission still occurs on multiple access channels. In terms of bandwidth, the analog solution is better than the digital solution solution (
M. M. Amiri \& G\"und\"uz, \citeyear{90amiri2020machine}). Digital solutions have the following advantages:
\begin{itemize}
\item Backward compatibility, they can be easily implemented on existing digital systems.
\item It is difficult to slow down users.
\item They are more reliable because various error control codes can be used.
\item Digital solutions do not require the tight synchronization required for analog transmission.
\end{itemize}

Driven by the above discussion, they considered the application of FL to multiple access channels. This study focuses on the design of a digital gradient transmission scheme, where the gradient of each user was the first quality conversion. This process is transmitted through multiple access channels and is decoded separately on the parameter server. The conditions are as follows: a) the informality of the gradient of each user, b) the underlying channel conditions, and proposed a stochastic gradient quantization scheme to optimize the quantization parameters according to the capacity area of the multiple access channel. The results show that especially when users experience different channel conditions or different degrees of information gradient, the channel-aware quantization of FL is better than the non-perceptual channel quantization scheme (for example, uniform distribution). The difference between this scheme and the scheme in (Suresh et al., \citeyear{92suresh2017distributed}) is that it allows each user to have its own quantitative budget. First, a scheme for an arbitrary user M is proposed, and the convergence speed of the scheme is analyzed. The algorithm proposes a general optimization problem of quantitative budget allocation based on multiple access channel capacities. Then, they showed an example with M = 2 users, and found the best quantitative budget and communication rate. To this end, they researched and analyzed a channel-aware quantization scheme that is superior to uniform quantization and other existing digital schemes.

\textcolor{blue}{\subsection{Challenges and problems}}

\noindent Current researches offer obvious opportunities from the edge to the core network. In Table \ref{tab:communication}, some studies have put forward certain solutions for the application of FL in communication. However, there are several key challenges related to the application of federated servers, as described in (Niknam et al., \citeyear{93niknam2020federated}).

Security and privacy. Although it adopts a secure aggregation algorithm, the encrypted local model can reveal the local situation by analyzing the global model. In the case of FL, the model was trained using sensitive user data. The premise of FL is to employ users to process data memory effectively without revealing private information. Ultimately, this process can reduce the potential for data disclosure in the event of an attack. Additionally, FL may be subject to reasoning and confrontational attacks. The enemy embeds carefully designed samples into the data, effectively affecting the local training datasets to manipulate the model-polluting FL process results. Therefore, it is necessary to explore how FL can improve its own defense mechanisms against these attacks.

Considerations such as the optimal number of local learners participating in the global update, the grouping of local learners, and the frequency of local updates and global aggregation that lead to a trade-off between model performance and resource protection, are all application-dependent and merit further study. Besides, for low-power devices such as IoT nodes, the scale of FL network updates may be enormous. So it is necessary to use sparse and compressed model parameters to reduce resource consumption. There have also been studies designed a FL-enabled intelligent fog radio access networks (F-RANs) based on accuracy correction and model compression (Z. Zhao et al., \citeyear{2020Federated}). A certainly feasible scheme is provided for the solution of this problem.

\begin{table}[htpb]
\centering
\caption{Category, Key contributions, and Framework in applications of FL in communication}
\label{tab:communication}
\resizebox{\textwidth}{!}{%
\begin{tabular}{|c|c|l|c|}
\hline
Category &
  Reference &
  \multicolumn{1}{c|}{Key contributions} &
  Framework \\ \hline
\multirow{4}{*}{\begin{tabular}[c]{@{}c@{}}\\communication\end{tabular}} &
  (Luping et al., \citeyear{89luping2019cmfl}) &
  \begin{tabular}[c]{@{}l@{}}
   1.Proposed a Communication-Mitigated Federated Learning (CMFL) framework.\\
   2.The communication overhead is greatly reduced, and the convergence of learning is ensured.\\
   3.The CMFL architecture improves communication efficiency and prediction accuracy.
  \end{tabular} &
  CMFL \\ \cline{2-4} 
 &
  (Amiri et al., \citeyear{91amiri2020federated}) &
  \begin{tabular}[c]{@{}l@{}}
  1.Proposed an analog communication scheme for compressed analog DSGD (CA-DSGD).\\
  2.Reduced bandwidth and outperformed other similar algorithms.\\
  3.Had faster convergence speed and higher accuracy.
  \end{tabular} &
  CA-DSGD \\ \cline{2-4} 
 &
  (Z. Zhao et al., \citeyear{2020Federated}) &
  \begin{tabular}[c]{@{}l@{}}
   1.Designed a federated learning enabled intelligent fog radio access networks (F-RANs).\\
   2.It provided a certain solution to the problem of data offload in the wireless network.\\
   3.Promoted network intelligent computing and reduced the high cost of model training.
  \end{tabular} &
  F-RANs 
  \\ \hline
\end{tabular}%
}
\end{table}

\textcolor{blue}{\section{The future development and direction of federated learning in smart cities}}

\noindent FL has been continuously developed since it was proposed in 2016. In addition to the main issues discussed at this stage (asynchronous (Li, Sahu,
Talwalkar, \& Smith, \citeyear{95li2020federated}), communication security, and privacy issues (Lim, Luong, et al., \citeyear{18lim2020federated})), the following key open directions remain to be explored.

Defense against attacks. Although FL can protect important information, if people deliberately launch a poisoning attack on distributed devices, it may also lead to the leakage of important information. For example, owing to the stochastic gradient descent (SGD) in the actual application process,  the leakage of these gradients may leak data information (Aono et al., \citeyear{100aono2017privacy}). Chuan et al. also studied the potential privacy and security issues in FL (C. Ma et al., \citeyear{75ma2020safeguarding}). Therefore, the ability to effectively defend against privacy and security issues in FL is an ongoing challenge.

Algorithm efficiency. The rapid growth of network traffic has become the main technical bottleneck in the development of IoT. Although FL can effectively connect distributed devices, optimization algorithms are also required to realize practical applications. For example, to reduce time complexity, the FedAvg algorithm (K. Yang et al., \citeyear{103yang2020federated}) is used for local calculation updates and aggregation, and is used for client-side differential privacy preservation federated optimization algorithm. Owing to the limitation of computing power, the related algorithms of FL still need to be optimized in the face of massive data.

Technology application. FL has widespread potential in smart city development. It can involve almost all aspects, especially in the fields of finance, medical care, transportation, etc. FL can be used to perform model training on data associated with multiple standards. Taking smart healthcare as an example, FL can train models that cannot be directly aggregated by hospitals. However, FL can fuse sensitive information without revealing privacy or overcoming the data island. Combining more data can significantly improve the accuracy of the model. The practical application of FL will also make cities smarter.

The gradual development of FL has introduced new opportunities to all walks of life. This article introduces the application of FL to smart cities, including communications, life services, and IoT. It is expected that in the near future, the use of FL will lead to the further development of smart cities. FL will also be combined with all walks of life to form a good ecological community, so that all persons can benefit from it.

\textcolor{blue}{\section{Conclusion}}

\noindent FL has been widely used and developed in various fields. This paper investigated the development achievements of FL in the fields of IoT, transportation, communication, medical care, and finance. Meanwhile, we consider the future research direction of FL in other fields related to smart cities. It will include heterogeneous communication security and privacy issues. We also consider proposing certain ideas and implementations in defense against attacks, potential privacy security, algorithm efficiency, and broader application scenarios. Moreover, we propose future prospects for technology development. Subsequently, we will continue to conduct in-depth research on key technologies. 

\textcolor{blue}{\section*{Acknowledgment}}

\noindent The authors would like to thank the anonymous reviewers for their constructive and insightful comments on this paper.

\textcolor{blue}{\section*{Disclosure statement}}

\noindent No potential conflict of interest was reported by the authors.

\textcolor{blue}{\section*{Funding}}

\noindent This work was supported by the Natural Science Foundation of Hainan Province under Grant No.619QN194.
\\

\bibliographystyle{apacite}
\bibliography{ref}

\end{document}